\documentclass{llncs}
\usepackage[latin1]{inputenc}
\usepackage{color}
\usepackage{longtable,multirow}
\usepackage[dvips]{graphicx} 
\usepackage{amsmath}
\usepackage{url}
\usepackage{algorithm,algorithmic}

\begin{document}


\title{KohonAnts: A Self-Organizing Ant Algorithm for Clustering and Pattern Classification}


\author{C. Fernandes\inst{1,2}, A.M. Mora\inst{2}, J.J. Merelo\inst{2}, V. Ramos\inst{1},J.L.J. Laredo\inst{2}}
\authorrunning{C. Fernandes et al.}

\institute{
LASEEB-ISR/IST. University of Lisbon (Portugal) \\
\email{\{cfernandes,vramos\}@laseeb.org}
\and 
Departamento de Arquitectura y Tecnología de Computadores. University of Granada (Spain) \\
\email{\{amorag,jmerelo,juanlu\}@geneura.ugr.es}
}

\maketitle
%
%
\begin{abstract}
In this paper we introduce a new ant-based method that takes advantage
of the cooperative self-organization of Ant Colony Systems to create a
naturally inspired clustering and pattern recognition
method. The approach considers each data item as an ant, which moves
inside a grid changing the cells it goes through, in a fashion similar
to Kohonen's Self-Organizing Maps. The resulting algorithm is
conceptually more simple, takes less free parameters than other ant-based clustering algorithms, and, after some parameter tuning, yields very good results on some benchmark problems.
\end{abstract}


%
%
\section{Introduction and State of the Art}
\label{sec:introproblema}
Clustering is performed naturally by ants at least in two different
ways. First, ant colonies recognize by odour other member of their
colony (as mentioned in the paper by \cite{labroche2003aac}) leading
to a natural clustering of ants belonging to the same nest, which is a consequence of nurturing and
also has some genetic support; second, ants do physically cluster
their larvae and dead bodies, putting them in piles whose position and
size is completely self-organizing, as described by \cite{deneubourg1991dcs}. 
Ant algorithms inspired by these models such as those proposed by 
\cite{bonabeau1998pok,abraham2003wum,labroche2003aac,vitor-AEB02} have been
applied to clustering and classification. In general, these methods
follow the second clustering behavior: data for training the clusters is represented as {\em dead bodies}, which ants have to pick up (with a certain
probability, and following some rule) and drop (also following some
rule), while at the same time dropping and following pheromones. This
results in the introduction of a few artifacts in the method: while
the number of {\em dead bodies} (data items) to sort is {\em natural},
grid size, number of ants, pheromone following behavior and the rest
is not. This results in a certain amount of parameter tuning for
obtaining good results, but in any case is farther away from natural inspiration. 

In this paper we present KohonAnts, an Ant Colony Optimization
algorithm that merges the biologically inspired concepts in Kohonen's
Self-Organizing Map (proposed and described in \cite{Kohonen88d,Kohonen2001}) and \cite{chialvomillonas95} ant algorithm (both will be introduced in next section). It is based in several new ideas. First, as in the above-mentioned Labroche et al. model, every ant represents a data item. Ants move in a grid dropping {\em vectorial} pheromones. The grid is filled with initially random vector pheromones (of the same dimension as the data), and every time an ant falls in a cell, it
changes the pheromone following a method similar to that used in
Kohonen Self-Organizing Map, making the cell pheromone closer to the
data item stored in the ant itself.

Since ants move around in the grid, ant position and pheromone content
co-adapt, so that eventually ants with similar data items are close
together in the grid (a {\em nesting} behavior), and the grid itself contains vectors similar to
those stored in the ants on top of them. The grid can then be used to
classify in the same way as Kohonen's Self-Organizing Map (but with better results), while ants can be used to visually identify the position of the clusters.

The interesting part of this method is that self-organization comes
through stigmergy: ants change their environment (pheromones stored on
the grid), and that influences the behavior of the rest of the ants
(that follow a path changed by their cluster-siblings). There are less
non-natural parameters (grid size is one of them), and, finally,
results obtained are quite competitive with other methods tested.

In this paper, after presenting all concepts used in our method in
section  \ref{sec:concepts}\textit{Preliminary Concepts}, after it, we will describe the KohonAnts model itself in section \ref{sec:model}\textit{Self-Organizing Ants Model}, followed by the experiments in section \ref{sec:experiments}\textit{Experiments and Results}. Finally, we will conclude our description in section \ref{sec:conclusions}\textit{Conclusions and Future Works} with a discussion of the obtained results and future lines of work.





\section{Preliminary Concepts}
\label{sec:concepts}

Before describing KohonAnts, we would like to introduce the algorithms in which 
it is based on for the unfamiliar reader. First, Ant Colony
Optimization algorithms are presented in subsection \ref{sec:aco}\textit{ACO}, followed by Kohonen's Self-Organizing Map in subsection \ref{sec:som}\textit{SOM}. Finally, Chialvo and Millonas' model is presented in subsection \ref{sec:asmodel}\textit{Ant System Model}.


\subsection{ACO}
\label{sec:aco}

The Ant Colony Optimization (ACO) is a meta-heuristic inspired by the
behavior of some species of ants that are able to find the shortest
path from nest to food sources in a short time. The method is based in the concept of \textit{stigmergy}, that is, communication between agents using the environment. Every ant, while walking, deposits a substance called \textit{pheromone} which other ants can sense. The ants tends to follow  pheromone (it evaporates after some time) so, in intersections between several trails, an ant moves with high probability following the highest pheromone level. 
This metaheuristic was introduced by Dorigo et al. in 1991 (see \cite{dori99a} and \cite{dori2002} for more details).

ACO algorithms take this behavior as inspiration to solve combinatorial optimization problems, using a colony of artificial ants as computational agents that communicate each other using {\em pheromones}. The problem to be solved using ACO must be transformed into a graph with weighted edges. In every iteration, each ant builds a complete path (solution), by travelling through the graph. At the end of this construction (and in some versions, during it), each ant leaves a trail in the visited edges depending on the fitness of the solution it has found. This is a measure of desirability for that edge and it will be considered by the following ants. In order to guide its movement, each ant uses two kinds of information that will be combined: \textit{pheromone trails},
which correspond to 'learnt information' changed during the algorithm
run, denoted by $\tau$; and \textit{heuristic knowledge}, which is a
measure of the desirability of moving to the next node, based in
previous knowledge about the problem (does not change during the
algorithm run), denoted by $\eta$. 
The ants usually choose edges with better values in both properties,
but sometimes they may 'explore' new zones in the graph because the
algorithm has a stochastic component, that broadens the search space
to regions not previously explored. Due to all these properties, all
ants cooperate in order to find the best solution for the problem
(the best path in the graph), resulting in an global emergent behavior.



There are lots of variants and new methods, but we introduce Ant Colony System (ACS) because our model takes some features of it.

The building of solutions is strongly based in the \textit{state transition rule} (called \textit{pseudo-random proportional state transition rule} in ACS), since every ant uses it to decide which node \textit{j} is the next in the construction of a solution (path), when the ant is at the node \textit{i}. This formula calculates the probability associated to every node in the neighbourhood of \textit{i}, and is as follows: 
\\
\begin{small}
If (\textit{q} $\le$ \textit{q}$_{0}$)
\begin{equation} \label{eq:acsrtj}
j=\arg \max  _{j\in N_{i} } \, \left\{ \sum\limits_{u\in N_{i}} \tau(i,u)^{\alpha}\cdot \eta(i,u)^{\beta}\right\} 
\end{equation}
\\
Else
\begin{equation} \label{eq:acsrtp}
P(i,j)=\left\{ \begin{array}{ll} 
\frac { \displaystyle \tau(i,j)^{\alpha}\cdot \eta(i,j)^{\beta}} 
      { \displaystyle \sum\limits_{u\in N_{i}} \tau(i,u)^{\alpha}\cdot \eta(i,u)^{\beta}} & \quad if \, j\in N_{i} \\
\\
\\
0 & \quad otherwise\\
\end{array} \right.
\end{equation}
\end{small}
Where \textit{q} is a random number in [0,1] and \textit{$q_0$} is a parameter which set the balance between exploration and exploitation. If $q \le q_0$, the best node is chosen as next (exploitation), on the other hand one of the feasible neighbours is selected, considering different probabilities for each one (exploration). \textit{$\alpha$} and \textit{$\beta$} are weighting parameters to set the relative importance of pheromone and heuristic information respectively, and \textit{N$_i$} is the current feasible neighbourhood for the node \textit{i}.

There is a \textit{global pheromone updating}, which is only performed for the edges of the global best solution, so for every edge $(i,j)$ in $S_{GlobalBest}$ is:
\begin{small}
\begin{equation} \label{eq:acsgfu}
\tau^{t}(i,j)=(1-\rho)\cdot \tau^{t-1}(i,j) + \rho \cdot \Delta\tau(i,j)_{GlobalBest}
\end{equation}
\end{small}
\textit{t} marks the new pheromone value and \textit{t-1} the old one.
\textit{$\rho$} in [0,1] is the common evaporation factor and \textit{$\Delta\tau$} is the amount of pheromone deposited depending on the quality of the best solution.

There is also a \textit{local pheromone updating}, which is performed by each ant, every time that a node \textit{j} is added to the path which it is building. This formula is:
\begin{small}
\begin{equation} \label{eq:acslfu}
\tau^{t}(i,j)=(1-\varphi)\cdot \tau^{t-1}(i,j) + \varphi \cdot \tau_0
\end{equation}
\end{small}
Where \textit{$\varphi$} in [0,1] is the local evaporation factor and \textit{$\tau_0$} is the initial amount of pheromone (it corresponds to a lower trail limit). 
This formula results in an additional exploration technique, because it
makes the edges traversed by an ant less attractive to the following
ants and helps to avoid that many ants follow the same path. 


\subsection{SOM}
\label{sec:som}

\begin{figure}[htpb]
\begin{center}
\includegraphics[scale=0.45]{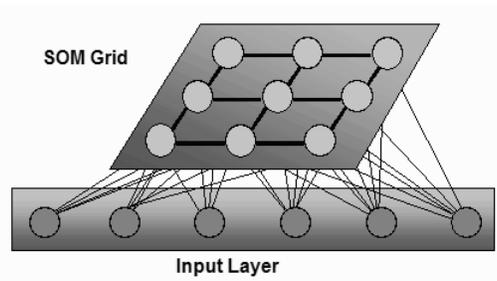}
\caption{SOM Grid
\label{fig:somgrid}}
\end{center}
\end{figure}

The Self-Organizing Map (SOM) was introduced by Teuvo Kohonen in
1982 (see \cite{Kohonen2001} for details). It is a non-supervised neural network that tries to imitate the self-organization done in the sensory cortex of
the human brain, where neighbouring neurons are activated by similar
stimulus. It is usually used either as a clustering/classification tool or as a method to find unknown relationships between a set of variables that
describe a problem. The main property of the SOM is that it makes a
nonlinear projection from a high-dimensional data space (one
dimension per variable) on a regular, low-dimensional (usually 2D)
grid of neurons (see Figure \ref{fig:somgrid}).

Since this type of network is distributed in a plane (2-dimensional structure) it can be concluded that the projections preserve the topologic relations while simultaneously creating a dimensional reduction of the representation space (the transformation is made in a topologically ordered way).

The SOM processes a set of input vectors (samples or patterns), which are composed by variables (features) typifying each sample, and creates an output topological network where each neuron is associated also to a vector of variables (model vector) which is representative of a group of the input vectors.
Note in Figure \ref{fig:somgrid} that each neuron of the network is completely connected to all the nodes (each node is a sample) of the input layer. So, the network represents a feed-forward structure with only one computational layer formed by neurons or model vectors.

There are four main steps in the processing of the SOM. Excepting the first one, the others are repeated until a stop criteria is reached:
\begin{itemize}
\item \textbf{Initialization of model vectors}. Usually it is made by assigning small random values to their variables, but there are some other possibilities as an initialization using random input samples.

\item \textbf{Competitive process}. For each input pattern $X$, all the neurons (model vectors) $V$ competes using a \textit{similarity function} in order to identify the most similar or close to the sample vector. The most usual function is a distance measure (as Euclidean distance). The winner neuron is called the best matching unit (BMU).

\item \textbf{Cooperative process}. The BMU determines the centre of a topological neighbourhood where those neurons inside it will be updated (the model vectors) to be even more similar to the input pattern. There is a \textit{neighbourhood function} used to determine the neurons to consider. If the lattice where the neurons are is rectangular or hexagonal, it is possible to consider as neighbourhood rectangles or hexagons with the BMU as centre. Although it is more usual to use a Gaussian function to assure that the farther the neighbour neuron is, the smaller the updating to its associated vector is.
In this process, the neurons inside a vicinity cooperate all of them to learn.

\item \textbf{Learning process}. In this step the variables of the model vectors inside the neighbourhood are updated to be closer to those of the input vector. It means doing the neuron more similar to the sample. The \textit{learning rule} used to update the vector ($V$) for every neuron $i$ in the neighbourhood of the BMU is:
{\small
\begin{equation}
\label{eq:learningrule}
V_i^{t} = V_i^{t-1} + \alpha^{t} \cdot N_{BMU}^{t}(i) \cdot (X-V_i^{t-1})
\end{equation}
}
Where t is the current iteration of the whole process, $X$ is the input vector, $N_{BMU}$ is the neighbourhood function for the BMU, which returns a high value (in [0,1]) if the neuron $i$ is in the neighbourhood and close to the BMU (1 if $i=BMU$), and a small value in the other case (0 if $i$ is not located inside the neighbourhood). $\alpha$ is the \textit{learning rate} (also in (0,1]).
Both (neighbourhood and learning rate) depends on $t$, since it is usual to decrease the radius of the first one and the value of the second in order to make higher updating at the beginning of the process and almost none in the latter.
\end{itemize}


The consecutive application of Equation \ref{eq:learningrule} and the
update of the neighbourhood function, has the effect of 'moving' the
model vectors, $V_j$ from the winning neuron towards the input vector
$X_i$. It is, the model vectors tend to follow the distribution of the
input vectors. Consequently, the algorithm leads to a topological
arrangement of the characteristic map of the input space, in the sense
that adjacent neurons in the network tend to have similar weights
vectors. 


As a consequence, looking at the display of a SOM, it is possible to recognize some clusters as well as the metric-topological relations of the data items (vectors of variables of the problem) and the outstanding variables.





\subsection{Ant System Model}
\label{sec:asmodel}

In \cite{chialvomillonas95}, the authors presented a simple
ant model where trails and networks of ant traffic emerge without
impositions by any special boundary conditions, lattice topology, or
additional behavioral rules. In this model, the state of an ant can
be expressed by its position $r$ and orientation $\theta$. Since the
response at a given time is assumed to be independent of the previous
history of the individual, it is sufficient to specify a transition
probability from one place and orientation ($r$, $\theta$) to the next
($r^*$, $\theta^*$) an instant later. Initial papers by \cite{millonas92,millonasalife3} transition rules were derived and generalized from noisy response functions, which in turn were found to reproduce a number of experimental results with real ants. The response function can effectively be translated into a two-parameter transition rule between the cells by using the pheromone weighting function showed in Equation \ref{eq:Wsigma}:
{\small
\begin{equation} 
\label{eq:Wsigma}
W(\sigma)= \left(
1 + \frac { \displaystyle \delta } 
          { \displaystyle 1 + \sigma \cdot \delta}
\right)^{\beta}
\end{equation}
}
This equation measures the relative probabilities of moving to a cell
$r$ with pheromone density $\sigma(r)$. The parameter $\beta$ is
associated with the osmotropotaxic sensitivity proposed in
\cite{wilsoninsects}. In practical terms, this parameter controls the
degree of randomness with which each ant follows the gradient of
pheromone: for low values of $\beta$, pheromone concentration does not
greatly affect its choice, while high values cause it to follow
pheromone gradient with more certainty, as proved in
\cite{chialvomillonas95}. The sensory capacity $1/\delta$ describes
the fact that each ant's ability to sense pheromone decreases somewhat
at high concentrations. In addition to the former equation, there is a
weighting factor $w(\Delta\theta)$, where $\Delta\theta$ is the change
in direction at each time step, i.e. measures the magnitude of the
difference in orientation. This weighting factor ensures that very
sharp turns are much less likely than turns through smaller angles;
thus each ant in the colony have a probabilistic bias in the forward
direction. A discretization of the model is necessary in order to
perform simulations and test some assumptions: Chialvo and Millonas
created a square lattice where ants can move around, taking one step at every iteration. The decision (where to go) is made according to the pheromone concentration in all  eight neighboring cells (Von Neumann neighborhood) and the weighting factor $w(\Delta\theta)$, using Equation \ref{eq:Wsigma}, and computing the transition probabilities via Equation \ref{eq:PikW}:
{\small
\begin{equation} 
\label{eq:PikW}
P_{ik} = \frac { W(\sigma_i) \cdot w(\Delta_i) } 
               { \sum\limits_{j/k} W(\sigma_j) \cdot w(\Delta_j) }
\end{equation}
}
This equation represents the transition probabilities on the lattice to go from cell $k$ to cell $i$ and notation $j/k$ indicates the sum over all the cells $j$ which are in the local (Von Neumann) neighborhood of $k$. $\Delta_i$ measures the magnitude of the difference in orientation for the previous direction at time $t-1$.
As an additional condition, each individual leaves a constant amount
$\eta$ of pheromone at the cell where it is located at every time step $t$. This pheromone decays at each time step at a rate $k$. Toroidal boundary conditions are imposed on the lattice to avoid  boundary effects. Please note that there is no direct communication between the organisms but a type of indirect communication through the pheromone field. In fact, ants are not allowed to have any memory and the individual's spatial knowledge is restricted to local information about the whole colony pheromone density.


This model has been applied in many different works, for instance in
\cite{vramosAnts2000}, the authors adapted it by placing the ants 'over' a gray-scale image. So, they evolve reinforcing pheromone levels around pixels with different gray levels yielding pheromone maps that may be a suitable support for edge detection and image segmentation.
This last model was improved in \cite{cfernandes05}  by introducing a mechanism to eliminate and create ants along the evolution process, which means a self-regulated population size and it results faster and also more effective in creating pheromone trails around the edges of the images.

\section{Self-Organizing Ants Model}
\label{sec:model}

The algorithm presented in this paper is an ant algorithm with some
common features with the Ant System of Chialvo et al., nevertheless it also
includes some other features inspired by the Kohonen's SOM. It is called, for
this reason, \textit{KohonAnts} (or KANTS). 

KANTS has been designed as a clustering and classification algorithm,
so it is capable to group a set of input samples (training dataset)
into clusters with similar features. In addition it behaves as a good
classification algorithm. It works in a non-supervised
(self-organizing) way, without considering the class of the input
patterns during the process. 

The main idea is to assign each input sample (which is a vector) to an
ant, and put them into an habitat which is a toroidal $X \cdot Y$
grid. Then, they move around in the lattice changing the environment,
which is a stigmergic mechanism. Every cell of the grid that
constitutes the environment also contains a vector of the same
dimension and range as the training set. The factor of change of the environment) depends on the values of the ant's vector, and, since every ant tends to move towards those zones in the grid which are more similar to themselves (to their associated vectors), ant position and pheromone content
co-adapt. This means that eventually, ants with similar data items
will be close together in the grid, and the grid itself will contain similar vectors to those stored in the ants on top of them. 

Then, the grid can be used as a classification tool (in the same way
as the resulting map after training using Kohonen's SOM), while ants will be grouped in clusters of similar individuals.

In the following paragraphs we present the most important features of the algorithm.




\subsection{{\em Decide Where to Go} Rule}

This is the most important function in the algorithm. It is used by
every ant placed at cell $i$ to decide which is the next cell $j$ to
move.

This function is based in Chialvo's Ants System pheromone weighting function and pseudo-random proportional rule of ACS, so it is:
\\
\begin{small}
If (\textit{q} $\le$ \textit{q}$_{0}$)
\begin{equation} \label{eq:kantsrtj}
j=\arg \max_{j\in N_{i}} \, W(\sigma_{ij}) 
\end{equation}
\\
Else
\begin{equation} \label{eq:kantsrtp}
P_{ij}=\left\{ \begin{array}{ll} 
\frac { \displaystyle W(\sigma_{ij})} 
      { \displaystyle \sum\limits_{u\in N^t_{i}} W(\sigma_{iu})} & \quad if \, j\in N^t_{i} \\
\\
0 & \quad otherwise\\
\end{array} \right.
\end{equation}
\end{small}
In that rule, \textit{q}$_{0}$ $\in$ [0,1] is the standard ACS parameter and \textit{q} is a random value in [0,1]. $N_i$ is the neighbourhood of the cell $i$, which is a function similar to the one used in SOM. It also has associated a \textit{neighbourhood radius, $nr$} which diminish along the running, so the neighbourhood is different at every iteration $t$. This function returns '1' if the cell is included in the neighbourhood and '0' otherwise.

$\sigma$ is defined by the following equation:
{\small
\begin{equation} 
\label{eq:kantssigma}
\sigma_{ij} =  \sqrt{V_i(v)^2 - CTR_j(v)^2} \quad\quad\forall v= 1..nvars
\end{equation}
}
Where $V_i$ is the vector associated to the cell $i$ and $CTR_j$ is the centroid of a zone centered in the cell $j$. It is a vector where each value takes the arithmetic mean of the correspondent values of the vectors associated to the cells included within a \textit{centroid radius, $cr$}. The formula is equivalent to calculate the Euclidean distance between the vector associated to the cell $i$ and the centroid vector for the cell $j$, both vectors have a number of variables $nvars$.

Finally, in the \textit{decide where to go} rule, $W(\sigma)$ is the Ant System pheromone weighting function (Equation \ref{eq:Wsigma}).

The rule works as follows: when an ant is building a solution path and is placed at one node $i$, a random number $q$ in [0,1] is generated, if $q \le q_{0}$ the best neighbour $j$ is selected as the next node in the path (Equation \ref{eq:kantsrtj}). Otherwise, the algorithm decides which node is the next by using a roulette wheel considering $P_{ij}$ as probability for every feasible neighbour $j$ (Equation \ref{eq:kantsrtp}).

Notice that the second part of the rule (Equation \ref{eq:kantsrtp}) is similar to the transition probability defined by Chialvo et al. (Equation \ref{eq:PikW}), but considering a weighting factor $w(\Delta\theta)=1$, so, all the neighbour cells have the same probability in advance (before considering the $\sigma$ value).

In addition, there is an important factor to mark, which is that the ants are capable to move to cells far more than one hop from the cell where they are currently located. It means that they can 'jump' or 'fly' as some real-world ant species are able. This property is vanishing along the algorithm running because the neighbourhood radius is decreased until it takes a value of '1' (ants only move from one cell to a one hop distance neighbour).


\subsection{The Updating Function}

This process is usually performed in classical ant algorithms as a pheromone trail deposition.
At every step, each ant $k$ updates the cell $i$ where is placed, using an updating formula similar to the learning function of SOMs (see Equation \ref{eq:learningrule}). Bearing in mind that every sample/ant and cell in the grid is a vector of $nvars$ variables, the formula is as follows:
{\small
\begin{equation} 
\label{eq:kantsupdate}
V_i^{t}(v) = V_i^{t-1}(v) + R \cdot [a_k(v) - V_i^{t-1}(v)] \quad\forall v= 1..nvars
\end{equation}
}
Where $V_i$ is the vector associated to the cell $i$, $t$ is the current iteration, and $a_k$ is the vector associated to the ant $k$.
$R$ is the reinforce of the update, which is described as:
{\small
\begin{equation} 
\label{eq:kantsreinforce}
R = \alpha \cdot (1-\overline{D}(a_k,CTR_i))
\end{equation}
}
$\alpha$ is the learning rate factor typical in SOM (which is constant in this algorithm), $CTR_i$ is again the centroid of a zone centered in the cell $i$. Finally, $\overline{D}$ is the mean Euclidean distance between the ant's vector and the centroid vector. It is:
{\small
\begin{equation} 
\label{eq:kantsmeandistance}
\overline{D} =  \sum\limits_{v=1}^{nvars} \frac{\sqrt{a_k(v)^2 - C_i(v)^2}}{nvars}
\end{equation}
}
%

\subsection{The Evaporation Function}

As in all the ant algorithms, it is a very important process where
the environment reverts to its previous (or initial) state. This
process is performed,for every cell $i$, once all the ants have moved
and updated the environment in the current iteration.
\begin{equation} 
\label{eq:kantsevaporate}
{\small
V_i(v) = V_i(v) - \rho \cdot V_{i0}(v) \quad\quad\forall v= 1..nvars
}
\end{equation}
Where $\rho$ is the usual evaporation factor and $V_{i0}$ is the initial vector associated to the cell $i$.
It means that the function changes the values of the vector in order to be similar to the initial, which can be interpreted as an evaporation of the trails in the environment.


\subsection{Pseudocode}

The pseudocode of our model is presented in Algorithms \ref{alg:kants} and \ref{alg:dwtg}. In these algorithms we consider each cell as a pair of coordinates, because is more accurate since the algorithm works using a grid.


\begin{algorithm} [ht]
\caption{KANTS Algorithm}
\label{alg:kants}
\begin{algorithmic}
\begin{small}
\STATE initialize\_randomly\_grid\_vectors
\STATE place\_randomly\_ants\_in\_grid
\FOR{N\_iterations}
	
	\FOR{each ant $a$ at cell $(x,y)$}
		\STATE $j$ = decide\_where\_to\_go($a$,$(x,y)$)
	\ENDFOR

	\STATE update\_grid    \quad// Using Equation \ref{eq:kantsupdate}
	\STATE evaporate\_grid    \quad// Using Equation \ref{eq:kantsevaporate}

	\STATE update\_neighbourhood\_radio
\ENDFOR
\end{small}
\end{algorithmic}
\end{algorithm}

\begin{algorithm} [ht]
\caption{Decide\_Where\_To\_Go ($a$,$(i,j)$)}
\label{alg:dwtg}
\begin{algorithmic}
\begin{small}
\FOR{all cells $(x,y)$ in neighbourhood of $(i,j)$}
   \STATE // Probability = Euclidean Distance to centroid
	\STATE $\sigma_{ij,xy}$ = E$_D$($(i,j)$,centroid($(x,y)$))
	\STATE compute $W(\sigma_{ij,xy})$ and $P_{ij,xy}$  \quad// Using Equations \ref{eq:Wsigma} and \ref{eq:kantsrtp}
\ENDFOR

\STATE // Ant Colony System/Ant System. Equations \ref{eq:kantsrtj} and \ref{eq:kantsrtp}
\STATE q = random(0,1)
\IF{q $\le$ q$_0$} 
   \STATE // selected cell = the one with maximum probability
	\STATE $(k,l)$ = MAX($P_{ij,xy}$)  
\ELSE
   \STATE // selected cell = roulette\_wheel
	\STATE $(k,l)$ = roulette\_wheel($P_{ij,xy}$)
\ENDIF
\end{small}
\end{algorithmic}
\end{algorithm}

%
%


\section{Experiments and Results}
\label{sec:experiments}

This section presents the data sets used to train and test KANTS algorithm (Subsection \ref{subsec:datasets}\textit{The Datasets}), followed by the results
obtained in clustering (Subsection \ref{subsec:expcluster}\textit{Clustering}) and classification (Subsection \ref{subsec:expclassif}\textit{Classification}).

\subsection{The Datasets}
\label{subsec:datasets}

The datasets used to test and validate the model are some well-known real world databases:

\begin{itemize}
\item\textbf{IRIS} contains data of 3 species of iris plant (Iris
Setosa, Versicolor  and Virginica), 50 samples of each one and
4 numerical attributes (the sepal and petal lengths and widths in
cms.). The first class is linearly separable from the others while the
other two are not. 
\item\textbf{GLASS} contains data from different types of glasses
studied in criminology. There are 6 classes, 214 samples (unevenly
distributed in classes) and 9 numerical features related to the
chemical composition of the glass. This database is difficult to
classify (and depending on the algorithm, also difficult to cluster),
since some classes are represented by just a few samples (3-10), and
some other classes not being linearly separable.
\item\textbf{PIMA}.
This is the Pima Indians Diabetes database which contains data related
to some patients (indians of that tribe) and  a class lebel
representing their diabetes diagnostic according to the world-wide health organization's criterion. There are 768 samples with 8 numerical features (medical data).
Again, this is a hard to process database, because many samples of the two classes takes close values for the same variables.
\end{itemize}

In each of the three databases, we have consider 3 sets built by transforming the original into 3 disjoint sets of equal size. The original class distribution (before partitioning) is maintained within each set. 
Then we consider 3 pair of datasets 'training-test' by splitting the 3 previous into half size ones, they are named including the text \textit{50tra-50tst}. In addition, 3 other pairs are created, but considering a distribution of 90\% of samples for training and 10\% for test. These sets are named including \textit{90tra-10tst}.



\subsection{Clustering}
\label{subsec:expcluster}

In \cite{chialvomillonas95}, the authors performed a study on the
distribution of ants with different configurations in the
$\beta$-$\delta$ parameter space. Three types of behavior were
observed when looking at the snapshots of the system after 1000
iterations: disorder, patches and trails.

The results obtained with their method follow theoretical prediction: a second order phase transition is observed, when a region of the parameter space which gives rise to disorder regimes "turns into" a region where trails are formed. Moving away from the order-disorder line, the system loses its ability to evolve lines/trails of ants and patches gradually appear. In addition, another experiment was conducted: the system was tuned to a region in the parameter space were trails emerge. After the traffic network was formed, $\beta$ was decreased in order to tune the system bellow the transition line; then, the ants started executing random walks and left their previously formed trails. Once $\beta$ was set again to the initial value, the ants self-organized again on a similar traffic network.

A similar test was performed with KANTS, but since Iris dataset was used (and due to it is not very complex), we have run the algorithm only a few iterations.
\begin{figure}[htp]
\begin{center}
\includegraphics[scale=0.6]{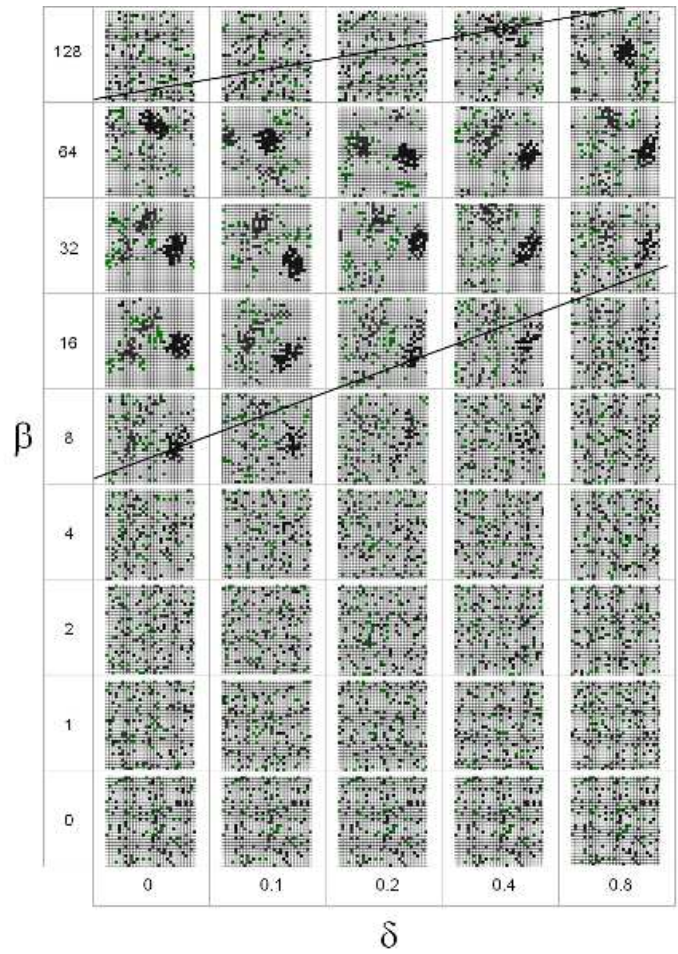}
\caption{Snapshots of the ants in the system after 100 iterations for
different $\beta$ and $\delta$ values. The straight lines roughly delimit the region where clusters emerge.
\label{fig:estudio-bd}}
\end{center}
\end{figure}

Parameters $\beta$ and $\delta$ were varied, and the resulting ants' distribution after 100 iterations is depicted in Figure \ref{fig:estudio-bd}. Parameters $\alpha$, \textit{neighbourhood radius (nr)} and \textit{centroid radius (cr)}, were set to 1, 1 and 3, respectively.
From the figures it is not possible to distinguish three different types of behavior, as in Chialvo and Millonas' experiments with the original model, but it is clear that there is a transition line from a disordered state, where ants/data do not cluster, and a ordered state where cluster start to emerge. Further away from the transition line, the model's ability to form clusters gradually starts to decay (again). In the same way as in the original model, there is only a small region of the parameter space that gives rise to a self-organized behavior, but while Ant System forms trails, KANTS emerge clusters of ants that are actually data samples.

Considering this results, KANTS appear to be a promising tool for data clustering. With a simple mechanism and proper tuning of $\beta$ and $\delta$, data represented by (and behaving as) ants form clusters that are easily distinguishable in the grid. Even if some kind of local search is eventually necessary in order to tackle real-world problems, KANTS by now come forward as a core model where hybridization may be performed and the resulting algorithms applied to hard problems.

\begin{figure}[h]
\begin{center}
\includegraphics[scale=0.3]{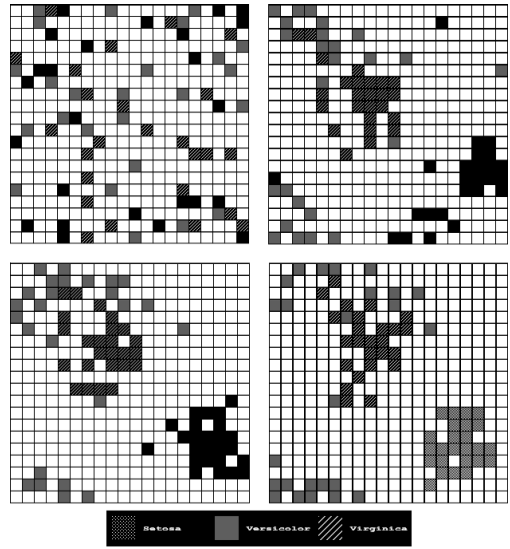}
\caption{Evolution of position of ants in the grid for the IRIS
  problem. It shows the situation at the beginning (top-left), at step
  50 (top-right) and 100 (bottom-left) and at step 150 (bottom-right).
\label{fig:evol_clusters}}
\end{center}
\end{figure}

In Figure \ref{fig:evol_clusters} an example of the ants evolution (movement during the run) in the grid is showed.

Looking at the snapshots of the grid at different iterations, it is possible to notice that every ant tends to move to a group of ants of the same class (they have similar values for the features). So, starting from a random initial configuration, in a few steps, the ants forms visible clusters.

\subsection{Classification}
\label{subsec:expclassif}

In order to classify with KANTS, we introduce a parameter: the number of neighbours to compare with the test sample. So, the algorithm searches for the $K$ nearest vectors in the grid (using the Euclidean distance) to the vector correspondent to the sample which it wants to classify. It assigns the class of the majority.

It is similar to the one used in K-Nearest Neighbours method (see \cite{KNN89} for details), but we use it once the grid has been trained (using the training dataset) and many times the algorithm works great even considering $K=1$.

Since KANTS is a stochastic approach, 10 runs were made considering each pair of datasets (training and test). Results are presented in Table \ref{tab:clasif_igp}, where mean, standard deviation, and best of the resulting percentages in classification are given.
We compare the results with those yielded using the traditional deterministic method K-Nearest Neighbours (KNN).

\begin{table} [htbp]
\centering
{\scriptsize
\begin{tabular}{|c|rr|rc|}

\hline
         IRIS  & \multicolumn{2}{|c}{{\it {\bf KANTS}}} & \multicolumn{ 2}{|c|}{{\it {\bf KNN}}} \\
\cline{2-5}
          Dataset & Best$\;$ & Mean$\quad\;\;$ & Best$\;$ & Mean \\
\hline
{\it 50tra-50tst-Set1} &      98.67 &      98.00 $\pm$0.67 &      97.30 &          - \\

{\it 50tra-50tst-Set2} &      98.67 &      97.60 $\pm$0.53 &      96.00 &          - \\

{\it 50tra-50tst-Set3} &     100.00 &      98.80 $\pm$0.40 &      94.60 &          - \\

{\it 90tra-10tst-Set1} &     100.00 &     100.00 $\pm$0.00 &     100.00 &          - \\

{\it 90tra-10tst-Set2} &     100.00 &      99.33 $\pm$2.00 &      93.33 &          - \\

{\it 90tra-10tst-Set3} &     100.00 &     100.00 $\pm$0.00 &      93.33 &          - \\
\hline
%
%
\hline
         GLASS & \multicolumn{2}{|c}{{\it {\bf KANTS}}} & \multicolumn{ 2}{|c|}{{\it {\bf KNN}}} \\
\cline{2-5}
          Dataset & Best$\;$ & Mean$\quad\;\;$ & Best$\;$ & Mean \\
\hline
{\it 50tra-50tst-Set1} &      68.22 &      65.42 $\pm$1.62 &      62.60 &          - \\

{\it 50tra-50tst-Set2} &      67.29 &      64.86 $\pm$1.52 &      64.40 &          - \\

{\it 50tra-50tst-Set3} &      74.77 &      71.03 $\pm$2.17 &      64.40 &          - \\

{\it 90tra-10tst-Set1} &      69.57 &     65.65 $\pm$1.30 &     47.80 &          - \\

{\it 90tra-10tst-Set2} &      73.91 &     73.48 $\pm$1.30 &      60.80 &          - \\

{\it 90tra-10tst-Set3} &      91.30 &     83.48 $\pm$3.25 &      82.60 &          - \\
\hline
%
%
\hline
         PIMA & \multicolumn{2}{|c}{{\it {\bf KANTS}}} & \multicolumn{ 2}{|c|}{{\it {\bf KNN}}} \\
\cline{2-5}
          Dataset & Best$\;$ & Mean$\quad\;\;$ & Best$\;$ & Mean \\
\hline
{\it 50tra-50tst-Set1} &      75.52 &      74.32 $\pm$0.61 &      70.03 &          - \\

{\it 50tra-50tst-Set2} &      77.34 &      76.61 $\pm$0.58 &      71.80 &          - \\

{\it 50tra-50tst-Set3} &      77.60 &      75.13 $\pm$0.85 &      72.90 &          - \\

{\it 90tra-10tst-Set1} &      83.12 &      80.52 $\pm$1.42 &      64.90 &          - \\

{\it 90tra-10tst-Set2} &      79.22 &      75.32 $\pm$1.42 &      73.60 &          - \\

{\it 90tra-10tst-Set3} &      84.42 &      80.65 $\pm$2.05 &      70.10 &          - \\
\hline
\end{tabular}
}
\caption{Classification results with Iris, Glass and Pima databases (6 different datasets each time).
\label{tab:clasif_igp} }
\end{table}

The results are very good when comparing them with a traditional
clustering and classification method such as KNN, even yielding 100\% in many
cases. We would like to enphasize the fact that the Glass and Pima datasets usually
obtain a low classification rate (both are difficult databases, as we
previously commented), while KANTS achieves in some cases a rate 10\%
higher than KNN. 

These results are remarkable since this is a non-supervised algorithm.

In addition is important to comment that the running time of the algorithm is just a few seconds, depending on the dataset size, so for these results it takes 8 seconds in Iris, 10 seconds in Glass and 20 seconds in Pima. All the experiments have been performed in a Pentium 1.6 GHz.


\section{Conclusions and Future Work}
\label{sec:conclusions}

This paper presents KohonAnts, a new method for clustering and data
classification, based on an hybridization of Ant Algorithms and Kohonen Self-Organizing Maps. The new model turns $n$-variable data samples into artificial ants that evolve in a 2D toroidal grid {\em paved} with $n$-dimensional vectors. Data/Ants act on the habitat vectors by pushing the values towards their own. In addition, ants are attracted by regions were the vector values are closer to their own data. In this way, similar ants tend
to aggregate in common regions of the grid. There is indirect
communication between ants through the grid (stigmergy) leading, with
a proper setting of the model's parameters, to the emergence of data
clusters. In addition, ants' actions (pheromone deposition) over the
grid and pheromone evaporation creates a kind of cognitive field which
has turned out be very effective for classification purposes.

It has been demonstrated that KANTS model is useful for
clustering and classification tasks, yielding very good results in
both kind of problems. The concept it is based on is quite simple and
naturally inspired, but even so results obtained are quite good
compared with traditional clustering methods (such as KNN). It is also a fast
method, not needing a lot of computation time for obtaining the
results mentioned above. As should be the spirit of publicly-funded
research, we maintain all sources for the project as well as data used
in experiments in the public repository {\scriptsize \url{https://forja.rediris.es/websvn/wsvn/geneura/KohonAnts/}}, under a GPL license\footnote{It is only requested that this paper (or another by the same authors) is referenced in published research.}.

As future short-term lines of work, we will perform further tests on the algorithm, comparing it with more specific clustering and classification methods. We will also try to streamline ant movement rules, and compare among different options. 

In addition, a lot of enhancements are still possible in the original KANTS model presented in this paper. A neighbourhood function may be considered, similar to the one used in Self-Organizing Maps for updating the environment in a radius.
As in \cite{cfernandes05} and in \cite{cfernandesICANN05}, reproduction may improve speed and accurateness of the algorithm. Chialvo and Millonas probability equation (Equation \ref{eq:PikW}) was not fully explored since weights $w(\Delta\theta)$ were set to 1.  Finally, a stopping criteria is needed in order to avoid unnecessary iterations in the process.

\section{Acknowledgements}

This work has been supported by NOHNES project from the Spanish
Ministry of Science and Education (TIN2007-68083-C02-01). C. Fernandes
also wishes to thank FCT, Ministério da Ciência e Tecnologia, his Research Fellowship SFRH/BD/18868/2004, also partially supported by Fundação para a Ciência e a Tecnologia (ISR/IST plurianual funding) through the POS\_Conhecimento Program that includes FEDER funds.

\footnotesize
\bibliographystyle{splncs}
\bibliography{KAnts}

\end{document}